\documentclass[journal]{IEEEtran}

\ifCLASSINFOpdf
\else
   \usepackage[dvips]{graphicx}
\fi
\usepackage{url}

\usepackage[numbers,sort&compress]{natbib}
\usepackage{graphicx}
\usepackage{makecell}
\usepackage[affil-it]{authblk}
\usepackage{stmaryrd}
\usepackage{amsbsy}
\usepackage{amsmath}
\usepackage{amssymb}
\usepackage{mathtools}
\usepackage{hyperref}
\usepackage{xcolor}

\begin{document}

\title{DART: Depth-Enhanced Accurate and Real-Time Background Matting}

\author[1,$\dagger$\thanks{$\dagger~$These authors contributed equally to this work.}]{Hanxi Li}
\author[1,$\dagger$]{Guofeng Li} 
\author[2,$\star$\thanks{$\star~$Corresponding author.}]{Bo Li} 
\author[3]{Lin Wu} 
\author[1]{Yan Cheng} 
\affil[1]{Jiangxi Normal University, Jiangxi, China}
\affil[2]{Northwestern Polytechnical University, Shaanxi, China}
\affil[3]{Swansea University, United Kingdom}

\markboth{Journal of \LaTeX\ Class Files, Vol. 14, No. 8, August 2015}
{Shell \MakeLowercase{\textit{et al.}}: Bare Demo of IEEEtran.cls for IEEE Journals}
\maketitle

\begin{abstract}
Matting with a static background, often referred to as ``Background Matting" (BGM), has garnered significant attention within the computer vision community due to its pivotal role in various practical applications like webcasting and photo editing. Nevertheless, achieving highly accurate background matting remains a formidable challenge, primarily owing to the limitations inherent in conventional RGB images. These limitations manifest in the form of susceptibility to varying lighting conditions and unforeseen shadows.

In this paper, we leverage the rich depth information provided by the RGB-Depth (RGB-D) cameras to enhance background matting performance in real-time, dubbed DART. Firstly, we adapt the original RGB-based BGM algorithm to incorporate depth information. The resulting model's output undergoes refinement through Bayesian inference, incorporating a background depth prior. The posterior prediction is then translated into a "trimap," which is subsequently fed into a state-of-the-art matting algorithm to generate more precise alpha mattes.
To ensure real-time matting capabilities, a critical requirement for many real-world applications, we distill the backbone of our model from a larger and more versatile BGM network. Our experiments demonstrate the superior performance of the proposed method. Moreover, thanks to the distillation operation, our method achieves a remarkable processing speed of 33 frames per second (fps) on a mid-range edge-computing device. This high efficiency underscores DART's immense potential for deployment in mobile applications \footnote{Source code are available at
\href{https://github.com/Fenghoo/DART}{https://github.com/Fenghoo/DART}.}

\end{abstract}


\IEEEpeerreviewmaketitle

\section{Introduction}

\IEEEPARstart{I}{mage} Image matting is a well-established problem in computer vision, with applications spanning image and video editing, video conferencing, and more. Researchers have dedicated substantial efforts to realizing automatic and high-quality image matting under real-world conditions, resulting in advancements such as deep learning-based approaches \cite{shen2016deep, zhu2017fast, chen2018semantic, liu2020boosting, sengupta2020background, li2022bridging, ke2022modnet, sun2022human, dai2022boosting, ma2023rethinking, li2023deep}. A significant challenge in automatic image matting lies in the requirement of a ``trimap," a crucial input for conventional matting algorithms. Without a proper trimap, the matting problem becomes ill-posed, as distinguishing between the ``foreground" and ``background" can be highly ambiguous \cite{li2023deep, xu2017deep, fang2022user, sun2023ultrahigh, park2022matteformer, liu2021long, zheng2022image, liu2021tripartite}. To address this issue, pioneering work such as the "Background Matting" (BGM) algorithm \cite{sengupta2020background, BGMv2} was introduced. BGM conducts image matting against a fixed background, ensuring a well-defined foreground that can be accurately extracted even without human-labeled trimaps. However, challenges persist, especially when dealing with shadows introduced by foreground objects or unexpected lighting variations. 

\begin{figure}[t!]  
  \centering
\includegraphics[width=0.45\textwidth]{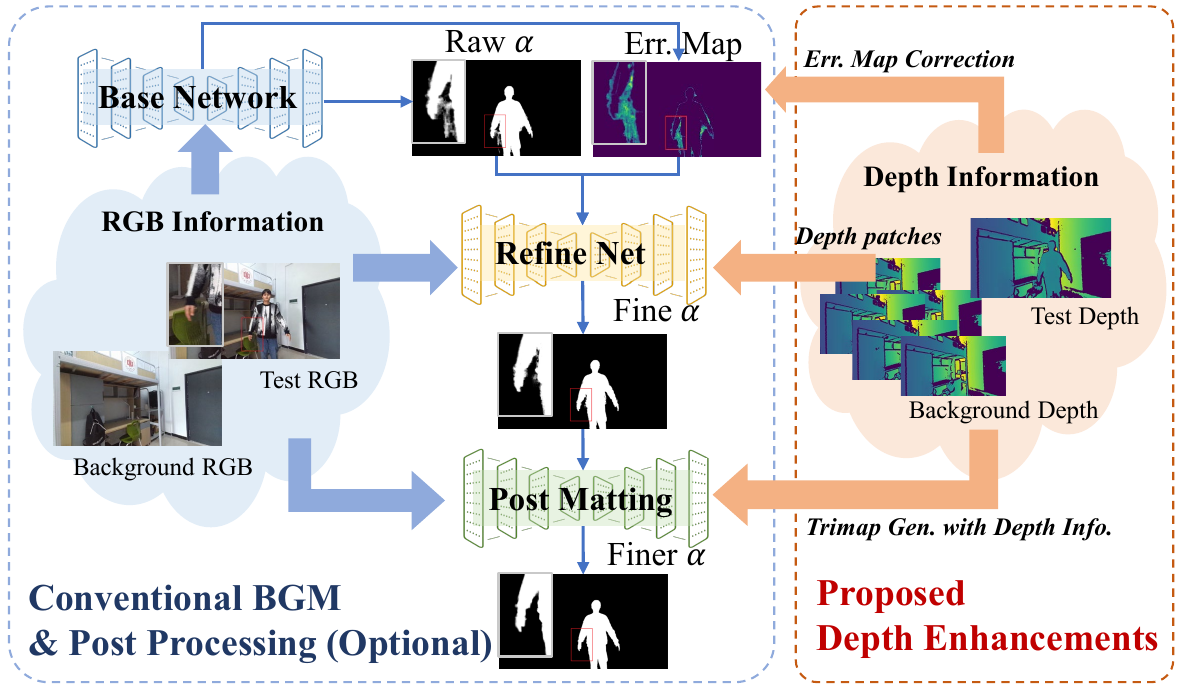}
  \caption{The illustration on the proposed depth-enhanced accurate and real-time backgournd matting (DART). Left: The conventional BGM and the optional post-matting process. Right: our depth-based enhancement approach.}
\label{fig:concept}
\end{figure}

In this paper, we present an innovative approach that leverages depth information obtained from a standard RGB-D camera to significantly improve the accuracy and robustness of image matting with static backgrounds. This novel utilization of depth data addresses some of the limitations inherent in traditional methods like BGM (Background Matting) and enhances the precision and reliability of matting results. We provide a schematic overview of our proposed method in Figure~\ref{fig:concept}. Our approach builds upon the foundation of the BGMv$2$ algorithm \cite{BGMv2} but introduces several key modifications to incorporate depth information. We refer to this method as ``Depth-enhanced Accurate and Real-Time BackGround Matting," or simply ``DART" for brevity. Specifically, we utilize depth information to correct the ``error map" estimated by the base network of the original BGM model \cite{BGMv2} and refine the ``trimap" used in the post-matting process. Additionally, we replace the use of traditional RGB patches with RGB-D patches in the refining network \cite{BGMv2}. Furthermore, we introduce the concept of distillation, where we derive a smaller base network \cite{hinton2015distilling} from the original ResNet$50$-based model used in BGMv$2$. These modifications collectively result in significantly improved matting performance compared to state-of-the-art methods, all while maintaining exceptional computational efficiency. Our DART algorithm can achieve a remarkable speed of $125$ FPS on an affordable desktop GPU and still maintains a respectable $33$ FPS on a mid-range edge-computing platform.

\section{The Proposed Method}

\begin{figure*}[htb!]  
\includegraphics[width=0.85\textwidth]{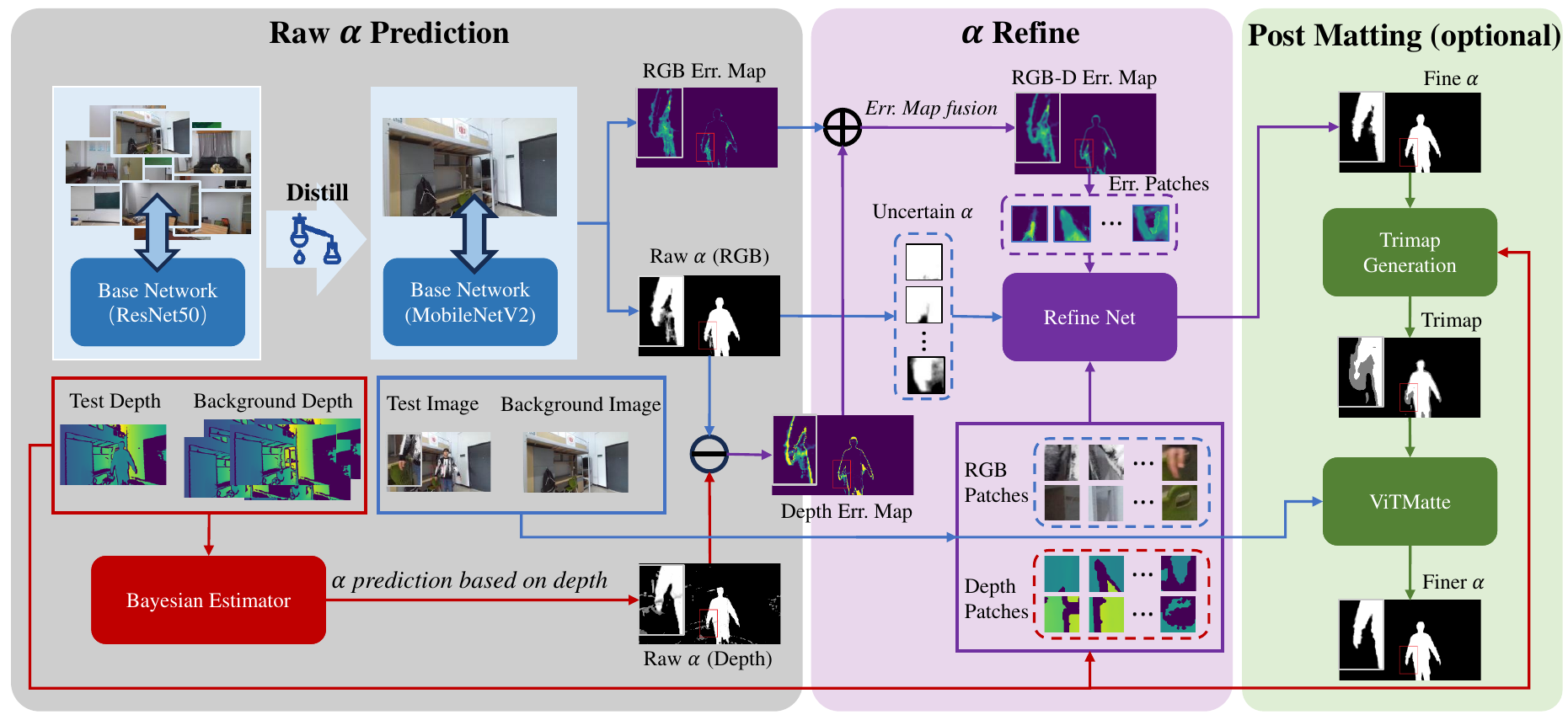}
  \centering
  \caption{The workflow of the proposed DART algorithm. The three stages of DART are shown
  in the gray, purple, and green regions. Better view in color.}
\label{fig:workflow}
\end{figure*}

The workflow of DART is illustrated in Fig.~\ref{fig:workflow}. As Fig.~\ref{fig:workflow}, we follow BGMv$2$ as the base network, and then we employ a smaller network (MobileNetv2
\cite{sandler2018mobilenetv2}) for higher efficiency and the network parameter is learned
via distilling the knowledge from the base network of BGMv$2$, followed by a fine-tuning on the current background. Besides, for different stages of the background matting process, the depth
information is exploited in different manners to increase the matting accuracy and
robustness. The remaining part of this section will detail the proposed depth-enhanced
BGM algorithm. 
  \vspace{-1.0em}
\subsection{Efficient base network from distillation}
In the original BGMv$2$ algorithm, the ResNet$50$-based \cite{he2016deep} ``base network'' of BGMv$2$ processes an input image ${\bf I} \in \mathbb{Z}^{H
\times W \times 3}$ as the following function: 
\begin{equation}
  \label{equ:resnet}
\Phi_{r}({\bf I}) \rightarrow \{{\bf A}^{\ast}_{raw}\in \mathbb{R}^{\frac{H}{4} \times
\frac{W}{4}}, {\bf E}^{\ast}_{RGB} \in \mathbb{R}^{\frac{H}{4} \times
\frac{W}{4}}\},
\end{equation}
where ${\bf A}^{\ast}_{raw} \in \mathbb{R}^{\frac{H}{4} \times \frac{W}{4}}$ denotes the raw
prediction of the alpha matte while ${\bf E}^{\ast}_{RGB}  \in \mathbb{R}^{\frac{H}{4} \times
\frac{W}{4}}$ is the predicted ``error map'' or in other words, the uncertain region of
${\bf A}^{\ast}_{raw}$ \cite{BGMv2}.

In this work, we employ a MobileNetv2-based \cite{sandler2018mobilenetv2} network to
predict the raw alpha of the test image. The similar inference process of this smaller model can be denoted as:
\begin{equation}
  \label{equ:mobile}
  \Phi_{m}({\bf I}) \rightarrow \{{\bf A}_{raw}\in \mathbb{R}^{\frac{H}{4} \times
  \frac{W}{4}}, {\bf E}_{RGB} \in \mathbb{R}^{\frac{H}{4} \times
  \frac{W}{4}}\}.
\end{equation}
where $\Phi_m({\bf I})$ denotes the MobileNetv2-based model which is learned via distillation. Following the SOTA distillation algorithm for segmentation \cite{shu2021channel}, the knowledge transfer from $\Phi_{r}$ to $\Phi_{m}$ can be realized via minimizing the following loss function:
\begin{equation}
  \label{equ:distill}
  \begin{split}
    L_{distill} =~ & KL({\bf A}_{raw}, {\bf A}^{\ast}_{raw})~ \\ 
    + ~ \|{\bf A}_{raw}& - {\bf A}_{GT}\|_{l_1} + \|{\bf E}_{RGB} - {\bf E}_{GT}\|_{l_2}, 
  \end{split}
\end{equation}
where ${\bf A}_{GT}$ and ${\bf E}_{GT}$ stand for the ground-truth alpha matte and error
map respectively; $KL(\cdot)$ denotes the Kullback–Leibler divergence between two
prediction maps. Note that in this paper we do not impose different weights for the three losses as the straightforward summation can already lead to sufficiently good results.  
  \vspace{-1.0em}
\subsection{Error map correction in a Bayesian style}
Besides the RGB images, the depth channel is also informative enough to estimate the
foreground region roughly. In this work, we conduct the depth-based matting in the
Bayesian manner. In particular, given the (resized) depth map of the test image ${\bf D}
\in \mathbb{R}^{\frac{H}{4} \times \frac{W}{4}}$ and the background depth map set
$\mathcal{D}_b = \{\forall~{\bf D}^i_b \in \mathbb{R}^{\frac{H}{4} \times \frac{W}{4}}
\mid i = 1, 2, \cdots, N\}$ \footnote{In this paper we assume that multiple static
background RGB-D images are captured before matting. Note that this is trivial to realize
considering that most RGB-D cameras can run faster than $30$ FPS.}, one can first fill
the unknown depth pixels with the special value $-1$. Then the mean value and the standard
deviation of the background depth on the coordinate $[r, c]$ can be calculated as: 
\begin{equation}
  \label{equ:mean_depth}
  \begin{split}
    \overline{\bf D}^{r, c}_{b} & = 
    \begin{cases}
      \frac{1}{|\mathcal{K}|}\sum_{i \in \mathcal{K}}{\bf D}^i_b(r, c), & |\mathcal{K}| > 0\\
      d_{max}, & |\mathcal{K}| = 0
    \end{cases} \\
    \sigma^{r,c}_b & = 
    \begin{cases}
      \sqrt{\frac{1}{|\mathcal{K}| - 1}\sum_{i \in \mathcal{K}}({\bf D}^i_b(r, c) -
      \overline{\bf D}^{r, c}_{b})^2}, & |\mathcal{K}| > 1\\
      \psi \cdot \overline{\bf D}^{r, c}_{b}, & |\mathcal{K}| = 1 \\
      \psi \cdot d_{max}, & |\mathcal{K}| = 0
    \end{cases}
  \end{split}
\end{equation}
where $d_{max}$ is the maximum detecting distance of the depth sensor; $\psi$ is a small
ratio; $\mathcal{K} = \{\forall i \mid {\bf D}^i_b(r, c) > 0 \}$ and $|\cdot|$ denotes the
set cardinality. Consequently, the conditional probability of an observed pixel depth $d$
on the coordinate $[r, c]$, given this pixel belongs to the foreground is calculated as:
\begin{equation}
  \label{equ:conditional_f}
    \text{P}^{r, c}_F(d) = \text{P}({\bf D}(r, c) = d \mid F) = 
  \begin{cases}
    \frac{1}{\overline{\bf D}^{r, c}_{b}}, & d \in (0, \overline{\bf D}^{r, c}_{b}]\\ 
    0, & \text{else}
  \end{cases} 
\end{equation}
Similarly, the conditional probability under the background condition writes:
\begin{equation}
  \label{equ:conditional_b}
    \text{P}^{r, c}_B(d) = \text{P}({\bf D}(r, c) = d \mid B) = 
  \begin{cases}
    \mathcal{N}^{+}_{r, c}(\overline{\bf D}^{r, c}_{b}, \sigma^{r, c}_b), & d > 0\\ 
    0, & \text{else}
  \end{cases}
\end{equation}
where $\mathcal{N}^{+}_{r, c}(\overline{\bf D}^{r, c}_{b}, \sigma^2_b)$ denotes the
estimated normal distribution of the background depth value on the pixel $[r, c]$
\footnote{Note that the negative part of the distribution is truncated and the probability
function is scaled so that the integral value over the domain $[0, \infty)$ is $1$.}.    

In a Bayesian way, the posterior probability that the pixel $[r, c]$ belongs to the foreground
can be calculated as:
\begin{equation}
  \label{equ:bayesian}
  \begin{split}
    \tilde{\text{P}}^{r, c}_F(d) = ~ & \text{P}(F \mid {\bf D}(r, c) = d) \\
    = ~ & \frac{\text{P}^{r, c}_F(d)\cdot \text{P}_F + \text{P}_F\cdot\zeta}{\text{P}^{r, c}_F(d)\cdot \text{P}_B +
    \text{P}^{r, c}_F(d)\cdot \text{P}_F + \zeta}
  \end{split}
\end{equation}
where $\text{P}_F$ and $\text{P}_B = 1 - \text{P}_F$ are the pre-defined prior probabilities of
foreground and background, respectively; $\zeta$ is a small-valued parameter to
ensure the foreground probability of a depth-unknown pixel equals $\text{P}_F$. We then
arrive at the foreground posterior map ${\bf A}_{D} \in \mathbb{R}^{\frac{H}{4} \times \frac{W}{4}}$ with
each element defined as: 
\begin{equation}
  \label{equ:post_depth}
  {\bf A}_{D}(r, c) = \tilde{\text{P}}^{r, c}_F({\bf D}(r, c)), ~\forall r, c.
\end{equation}
In practice, we post-process ${\bf A}_{D}(r, c)$ by gaussian blurring and small region
removal for more robust prediction. 

Recall that the RGB-based raw alpha matte defined in Equ.~\ref{equ:mobile}, we can
compare the two foreground probability maps to generate a residual map as 
\begin{equation}
  \label{equ:alpha_comp}
  {\bf E}_D = |{\bf A}_{D} - {\bf A}_{raw}|^{\circ} \in \mathbb{R}^{\frac{H}{4} \times \frac{W}{4}},
\end{equation}
where $|\cdot|^{\circ}$ denotes the element-wise absolute value. The residual map is used
as a complement to the RGB-based error map ${\bf E}_{RGB}$ and the corrected error map is
given by:
\begin{equation}
  \label{equ:corrected_err}
  {\bf E}_{RGBD} = \beta \cdot {\bf E}_{D} + (1 - \beta) \cdot {\bf E}_{RGB}, 
\end{equation}
where $\beta$ is the balancing parameter for fusing the two error maps.

\subsection{Alpha refinement with RGB-D patches}
The second stage of the conventional BGM algorithm is the patch-level correction to the
raw prediction. In the RGB-D scenario of this work, we propose to employ the RGB-D patch,
rather than the RGB path as the input of the refinement net. Considering that the depth
information is relatively independent to the RGB content, the RGB-D patch could lead to
a more accurate correction. This benefit is also proved empirically in the experiment part. 
In a mathematical way, the refining model in this work can be defined as 
\begin{equation}
  \label{equ:refine}
  \Omega({\bf I}, {\bf D}, {\bf A}_{raw}, {\bf E}_{RGBD}) \rightarrow {\bf A}_{fine} \in
  \mathbb{R}^{H \times W}
\end{equation}
\subsection{Post matting with depth information}
Given the sufficient computational or time budget, the optional post-matting can
significantly increase the matting accuracy. In this paper, we employ the depth
information to generate a better ``trimap'' for the SOTA matting algorithm
\cite{vitmatte}. Recall that the refined alpha prediction is denoted as ${\bf A}_{fine}$
which can be viewed as the ``prior'' foreground probability to the next inference stage,
thus the posterior map considering the depth evidence is calculated as:
\begin{equation}
  \label{equ:post_finer}
    \tilde{{\bf A}}^{r, c}_{fine} = ~ \frac{\text{P}^{r, c}_F(d)\cdot {\bf
    A}^{r,c}_{fine}}{\text{P}^{r, c}_F(d)\cdot(1 - {\bf A}^{r,c}_{fine}) + \text{P}^{r, c}_F(d)\cdot
    {\bf A}^{r,c}_{fine}}.
\end{equation}
We then generate a trimap based on the posterior map $\tilde{{\bf A}}_{fine}$ as
\begin{equation}
  \label{equ:trimap}
  \tilde{{\bf A}}_{fine} \xrightarrow{\text{Gaussian Blur}} \tilde{{\bf
  A}}^{\dagger}_{fine} \xrightarrow{\text{Two Threshoulds}} {\bf T} \in
  \mathbb{R}^{H \times W}, 
\end{equation}
where $\bf T$ is a triple-valued map with $1$-valued pixels indicating the foreground
objects, $0$-valued ones indicating the background area and $0.5$-valued ones standing for
the unknown region. This trimap is finally fed into the SOTA ViTMatte algorithm
\cite{vitmatte} for generating more accurate foreground masks. Note that this post-matting process is optional as the better performance is achieved at the cost of speed decreasing. 

\subsection{Proposed Dataset}
\begin{figure}[htb!]  
  \centering
  \includegraphics[width=0.45\textwidth]{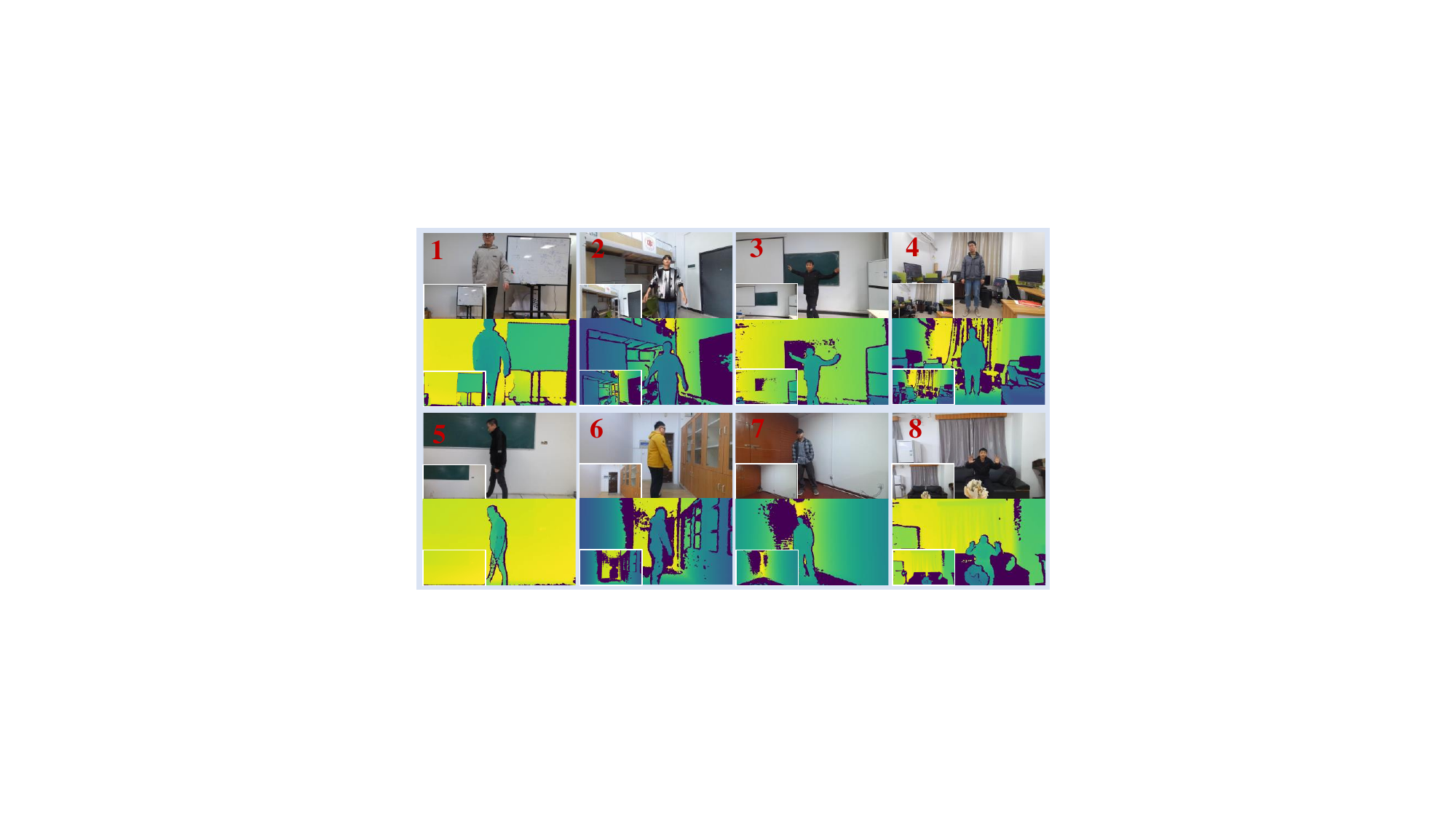}
  \caption{The proposed RGB-D background matting dataset. $8$ scenes are illustrated here,
  each with one RGB image pair (test and background) and one depth map pair. The dark blue
  pixels of the depth map stand for the depth-unknown region. Better view in color.}
\label{fig:dataset}
\end{figure}

As we introduced before, no RGB-D background matting dataset is available so far. We
thus design and produce a specific dataset for this task. The proposed dataset, termed
''JXNU RGBD Background Matting'' or JXNU-RGBD for short, contains
$12$ indoor scenes, with each one involves $100$ pure background RGB-D images and $5$
RGB-D images with foreground objects for testing. Only the alpha matte on the test images
is manually labeled for algorithm evaluation, and the labels are strictly unseen during
training.  Fig.~\ref{fig:dataset} illustrates $8$ out of the $12$ scenes of the proposed
dataset.
  \vspace{-1.0em}
\subsection{Training strategy and implementation detail}
Following the standard training protocol of background matting \cite{BGMv2}, we train our
DART model only based on real background (RGB-D) images and synthetic (RGB-D) foreground
objects. In particular, the ResNet$50$-based base network is trained based on the training
set proposed in BGMv$2$ \cite{BGMv2} augmented with the foreground objects (RGB)
extracted from the X-Humans dataset \cite{shen2023x}. Secondly, the ResNet$50$ model is then
fine-tuned on the proposed JXNU-RGBD dataset with only background images (RGB) and
artificial foregrounds. Finally, the MobileNetv2-based base network is distilled from the
fine-tuned larger model based upon the background images (RGB) merely from the target
scene. As to the refine net of DART, we first fuse the rendered RGB-D samples from the
X-Humans dataset \cite{shen2023x} with all the background RGB-D images of JXNU-RGBD to
train the raw model and then slightly fine-tune it using only the information from the
target scene. 

As to the hyperparameters, we set $\beta = 0.05$ to fuse the two error map, $\psi = 0.01$,
$d_{max} = 5460$, $\zeta = 0.001$. The two thresholds for generating trimap are $[0.25,
0.8]$. When distilling the base network, the batch size is set to $16$ and the learning
rates for the three submodules of the base network are $[1e-4, 5e-4, 5e-4]$. 

 \vspace{-1.0em}

\section{Experiment}
\label{sec:exp}

\subsection{Basic settings}
\label{subsec:basic}
In this section, we perform a series of experiments to evaluate the proposed method,
compared with the BGM-V$2$ \cite{BGMv2} (the SOTA background matting method); ViTMatte
\cite{vitmatte} (the SOTA general purpose matting algorithm); HIM \cite{sun2022human},
SGHM \cite{Chen_2022_ACCV} and P3M-Net \cite{ma2023rethinking} (three SOTA human matting
algorithms).

Four commonly used matting metrics, namely the sum of absolution difference (SAD), mean
square error(MSE), gradient error (Grad), and connectivity error (Conn) are employed for
scoring the comparing methods. We also report the FPS numbers for each matting algorithm.
Most experiments are conducted on a desktop PC with an Intel i5-13490F CPU, 32G DDR4 RAM
, and an NVIDIA RTX4070Ti GPU. To evaluate the compatibility of the proposed method on the
edge-computing devices, we also run the proposed method on an NVIDIA Jetson Orin NX
development board, with the reimplemented code employing the TensorRT SDK \cite{migacz20178} and the FP16
data format.     
  \vspace{-1.0em}
\subsection{Quantitative Results}
The matting accuracies of the involved methods are reported in Tab.~\ref{tab:result},
along with the corresponding running FPS numbers. Note that ViTMatte requires a trimap as
a part of its input, we thus generate a trimap for it based on the posterior map defined
in Equ.~\ref{equ:bayesian}.   
  \vspace{-1.0em}
\begin{table}[htb]
\label{tab:result}
\centering
\caption{The matting performances of the proposed method and the comparing SOTA algorithms.
  The running FPS of each comparing algorithm is also illustrated in the last
  column. Note that MSE is divided by $1000$ for ease of reading. The speed is measured on
  the desktop environment described in Sec.~\ref{subsec:basic}.
  }
 \resizebox{\columnwidth}{!}
  {
  \begin{tabular}{ l | c | c | c | c | c }
	\hline
  Methods                               & SAD~$\downarrow$ & MSE~$\downarrow$ &
    Grad.~$\downarrow$ & Conn$\downarrow$ & FPS ~$\uparrow$~ \\
		\hline

  VITMatte\cite{vitmatte}      & $17.71$               & $11.16$           & $21.67$     & $16.60$               & $5$           \\
   P3M-Net \cite{ma2023rethinking}       & $18.78$               & $8.60$           & $10.9$               & $18.57$               & $4$            \\ 
   SGHM \cite{Chen_2022_ACCV}             & $6.95$                & $2.67$             & $8.51$                 & $6.41$                 & $12$           \\ 
     HIM \cite{sun2022human}               & $4.28$               & $\textcolor{blue}{\underline{1.09}}$   & $ \textcolor{blue}{\underline{6.92}}$               & $3.55$               &  $4$       \\   BGM \cite{BGMv2}                      & $4.78$               & $1.86$           &$10.05$               & $4.67$               & $\textcolor{blue}{\underline{81}}$           \\
       DART  & $\textcolor{blue}{\underline{3.39}}$    & $1.22$            & $8.89$              & $\textcolor{blue}{\underline{3.33}}$ & $\textcolor{red}{\bf 125}$ \\
    DART  + ViTMatte\cite{vitmatte}       & $\textcolor{red}{\bf 2.90}$    & $\textcolor{red}{\bf 0.61}$        & $\textcolor{red}{\bf 6.02}$        & $\textcolor{red}{\bf 2.42}$  & $5$           \\
    \hline
  \end{tabular}
}
\end{table}

According to the table, one can clearly see the superiority of the proposed DART
algorithm. In particular, the DART with post-processing beats all the SOTA methods
evaluated by all four metrics; the vanilla DART algorithm achieves slightly worse matting
accuracy compared with its sophisticated version but enjoys a $25$-time faster speed.  

\begin{figure}[htb!]  
  \centering
  \includegraphics[width=0.45\textwidth]{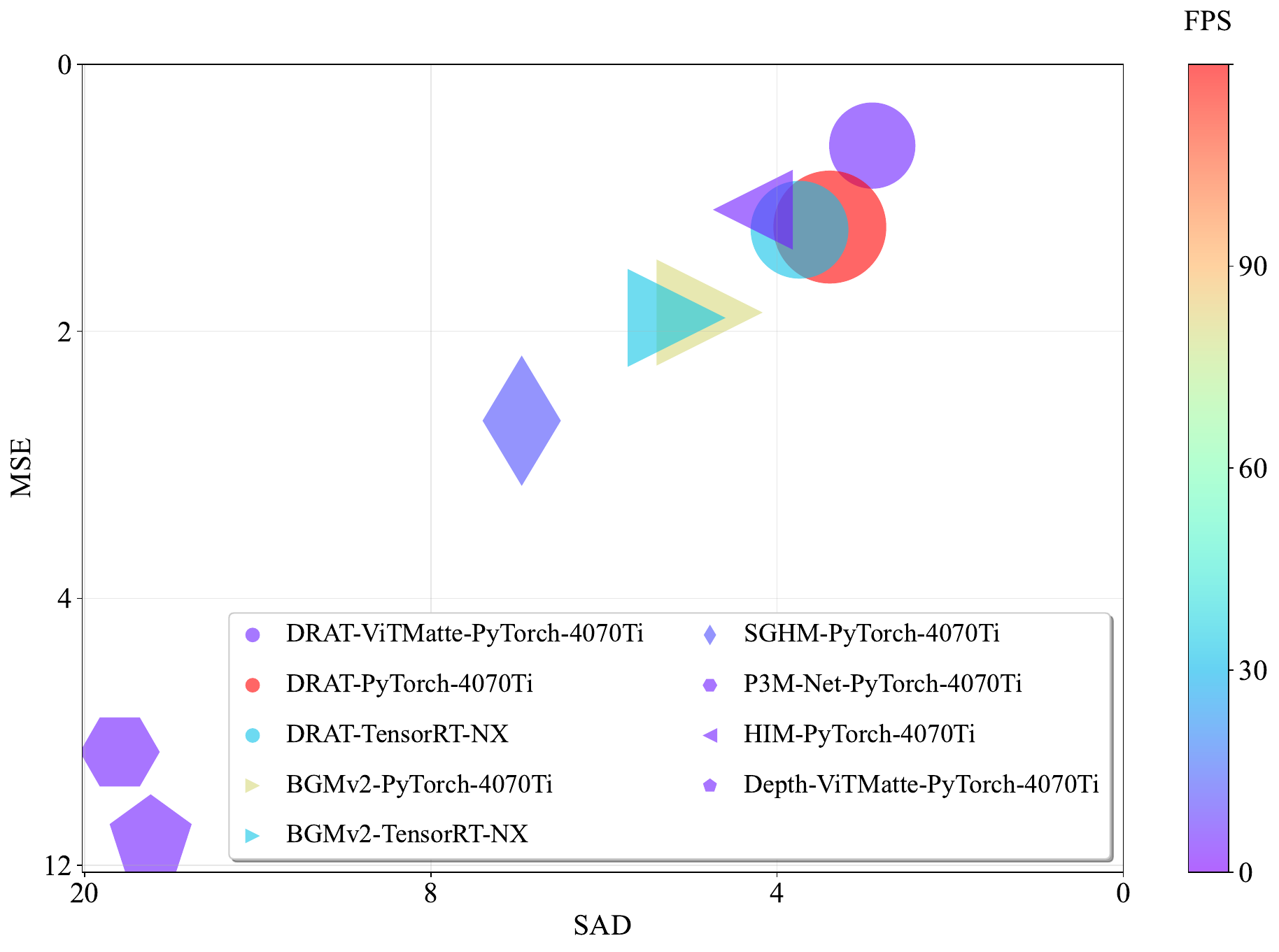}
  \caption{Speed and accuracy comparison of involved matting methods. Better view in
  color.}
\label{fig:speed}
\end{figure}

Fig.~\ref{fig:speed} demonstrates the performances of the compared methods from another
angle. The marker's location indicates the matting accuracy while the algorithm speed is
represented by the marker's color (the redder, the faster) and size (the larger, the
faster). From the figure, we can see the proposed DART algorithm performs fastest among all
the comparing methods. On the edge-computing platform (NVIDIA Jetson Orin NX), our method
can also achieve real-time speed ($33$ FPS) while still enjoying a remarkable accuracy
advantage to the fast BGM-V$2$ algorithm \cite{BGMv2}.    
  \vspace{-1.0em}
  \vspace{-1.0em}
\subsection{Ablation study}
The ablation study is shown in Tab.~\ref{tab:ablation} from where we can see a consistent
increasing trend of matting error when the proposed modules are removed from the DART
algorithm one by one.

\begin{table}[htb]
  \vspace{-1.0em}
\centering
\caption{The ablation study of the proposed method.}
\label{tab:ablation}
 \resizebox{\columnwidth}{!}
  {
  \begin{tabular}{ l | c | c | c | c  }
		\hline
  Methods                      & SAD~$\downarrow$ & MSE~$\downarrow$ & Grad~$\downarrow$ & Conn$\downarrow$  \\
		\hline
  DART + ViTMatte\cite{vitmatte}         & $2.9$           & $0.61$          & $6.02$   & $2.42$            \\
  DART                                   & $3.39$          & $1.22$          & $8.89$   & $3.33$             \\
  DART-NoErrorMap                        & $3.56$          & $1.26$          & $9.09$   & $3.45$             \\
  DART-NoDepthRefine                     & $4.07$          & $1.75$          & $10.31$  & $3.75$             \\
  DART-NoErrorMap-NoDepthRefine          & $4.27$          & $1.8$           & $10.43$  & $3.92$              \\
		\hline
  \end{tabular}
}
  \vspace{-1.0em}
\end{table}


\section{Conclusion}

This paper introduces a fixed-background matting algorithm that is enhanced by depth
information. By smartly exploiting the useful and complementary depth channel of an RGB-D
image, the proposed DART algorithm achieves more accurate results compared with the
existing SOTA matting approaches. Meanwhile, thanks to the successful distillation
process, our method is fast: it runs at $125$ FPS on a GPU-equipped desktop PC and $33$
FPS on a mid-range edge-computing platform. To evaluate the proposed algorithm, we make a
dedicated dataset, termed ``JXNU-RGBD'' for the RGB-D background matting task.  To the best of our
knowledge, this paper is the first work in the literature that explores the RGB-D
matting problem with a fixed background. It paves the way to future better solutions for
this new but realistic computer vision problem. 

\bibliography{SPL_DART}

\begin{thebibliography}{10}
\providecommand{\url}[1]{#1}
\csname url@samestyle\endcsname
\providecommand{\newblock}{\relax}
\providecommand{\bibinfo}[2]{#2}
\providecommand{\BIBentrySTDinterwordspacing}{\spaceskip=0pt\relax}
\providecommand{\BIBentryALTinterwordstretchfactor}{4}
\providecommand{\BIBentryALTinterwordspacing}{\spaceskip=\fontdimen2\font plus
\BIBentryALTinterwordstretchfactor\fontdimen3\font minus \fontdimen4\font\relax}
\providecommand{\BIBforeignlanguage}[2]{{%
\expandafter\ifx\csname l@#1\endcsname\relax
\typeout{** WARNING: IEEEtran.bst: No hyphenation pattern has been}%
\typeout{** loaded for the language `#1'. Using the pattern for}%
\typeout{** the default language instead.}%
\else
\language=\csname l@#1\endcsname
\fi
#2}}
\providecommand{\BIBdecl}{\relax}
\BIBdecl

\bibitem{shen2016deep}
X.~Shen, X.~Tao, H.~Gao, C.~Zhou, and J.~Jia, ``Deep automatic portrait matting,'' in \emph{European Conference on Computer Vision}.\hskip 1em plus 0.5em minus 0.4em\relax Springer, 2016, pp. 92--107.

\bibitem{zhu2017fast}
B.~Zhu, Y.~Chen, J.~Wang, S.~Liu, B.~Zhang, and M.~Tang, ``Fast deep matting for portrait animation on mobile phone,'' in \emph{Proceedings of the 25th ACM international conference on Multimedia}, 2017, pp. 297--305.

\bibitem{chen2018semantic}
Q.~Chen, T.~Ge, Y.~Xu, Z.~Zhang, X.~Yang, and K.~Gai, ``Semantic human matting,'' in \emph{Proceedings of the 26th ACM international conference on Multimedia}, 2018, pp. 618--626.

\bibitem{liu2020boosting}
J.~Liu, Y.~Yao, W.~Hou, M.~Cui, X.~Xie, C.~Zhang, and X.-s. Hua, ``Boosting semantic human matting with coarse annotations,'' in \emph{Proceedings of the IEEE/CVF Conference on Computer Vision and Pattern Recognition}, 2020, pp. 8563--8572.

\bibitem{sengupta2020background}
S.~Sengupta, V.~Jayaram, B.~Curless, S.~M. Seitz, and I.~Kemelmacher-Shlizerman, ``Background matting: The world is your green screen,'' in \emph{Proceedings of the IEEE/CVF Conference on Computer Vision and Pattern Recognition}, 2020, pp. 2291--2300.

\bibitem{li2022bridging}
J.~Li, J.~Zhang, S.~J. Maybank, and D.~Tao, ``Bridging composite and real: towards end-to-end deep image matting,'' \emph{International Journal of Computer Vision}, pp. 246--266, 2022.

\bibitem{ke2022modnet}
Z.~Ke, J.~Sun, K.~Li, Q.~Yan, and R.~W. Lau, ``Modnet: Real-time trimap-free portrait matting via objective decomposition,'' in \emph{Proceedings of the AAAI Conference on Artificial Intelligence}, 2022, pp. 1140--1147.

\bibitem{sun2022human}
Y.~Sun, C.-K. Tang, and Y.-W. Tai, ``Human instance matting via mutual guidance and multi-instance refinement,'' in \emph{Proceedings of the IEEE/CVF Conference on Computer Vision and Pattern Recognition}, 2022, pp. 2647--2656.

\bibitem{dai2022boosting}
Y.~Dai, B.~Price, H.~Zhang, and C.~Shen, ``Boosting robustness of image matting with context assembling and strong data augmentation,'' in \emph{Proceedings of the IEEE/CVF Conference on Computer Vision and Pattern Recognition}, 2022, pp. 11\,707--11\,716.

\bibitem{ma2023rethinking}
S.~Ma, J.~Li, J.~Zhang, H.~Zhang, and D.~Tao, ``Rethinking portrait matting with privacy preserving,'' \emph{International journal of computer vision}, pp. 1--26, 2023.

\bibitem{li2023deep}
J.~Li, J.~Zhang, and D.~Tao, ``Deep image matting: A comprehensive survey,'' \emph{arXiv preprint arXiv:2304.04672}, 2023.

\bibitem{xu2017deep}
N.~Xu, B.~Price, S.~Cohen, and T.~Huang, ``Deep image matting,'' in \emph{Proceedings of the IEEE conference on computer vision and pattern recognition}, 2017, pp. 2970--2979.

\bibitem{fang2022user}
X.~Fang, S.-H. Zhang, T.~Chen, X.~Wu, A.~Shamir, and S.-M. Hu, ``User-guided deep human image matting using arbitrary trimaps,'' \emph{IEEE Transactions on Image Processing}, pp. 2040--2052, 2022.

\bibitem{sun2023ultrahigh}
Y.~Sun, C.-K. Tang, and Y.-W. Tai, ``Ultrahigh resolution image/video matting with spatio-temporal sparsity,'' in \emph{Proceedings of the IEEE/CVF Conference on Computer Vision and Pattern Recognition}, 2023, pp. 14\,112--14\,121.

\bibitem{park2022matteformer}
G.~Park, S.~Son, J.~Yoo, S.~Kim, and N.~Kwak, ``Matteformer: Transformer-based image matting via prior-tokens,'' in \emph{Proceedings of the IEEE/CVF Conference on Computer Vision and Pattern Recognition}, 2022, pp. 11\,696--11\,706.

\bibitem{liu2021long}
Q.~Liu, H.~Xie, S.~Zhang, B.~Zhong, and R.~Ji, ``Long-range feature propagating for natural image matting,'' in \emph{Proceedings of the 29th ACM International Conference on Multimedia}, 2021, pp. 526--534.

\bibitem{zheng2022image}
Y.~Zheng, Y.~Yang, T.~Che, S.~Hou, W.~Huang, Y.~Gao, and P.~Tan, ``Image matting with deep gaussian process,'' \emph{IEEE Transactions on Neural Networks and Learning Systems}, 2022.

\bibitem{liu2021tripartite}
Y.~Liu, J.~Xie, X.~Shi, Y.~Qiao, Y.~Huang, Y.~Tang, and X.~Yang, ``Tripartite information mining and integration for image matting,'' in \emph{Proceedings of the IEEE/CVF International Conference on Computer Vision}, 2021, pp. 7555--7564.

\bibitem{BGMv2}
S.~Lin, A.~Ryabtsev, S.~Sengupta, B.~L. Curless, S.~M. Seitz, and I.~Kemelmacher-Shlizerman, ``Real-time high-resolution background matting,'' in \emph{Proceedings of the IEEE/CVF Conference on Computer Vision and Pattern Recognition}, 2021, pp. 8762--8771.

\bibitem{hinton2015distilling}
G.~Hinton, O.~Vinyals, and J.~Dean, ``Distilling the knowledge in a neural network,'' in \emph{NIPS Deep Learning and Representation Learning Workshop}, 2015.

\bibitem{sandler2018mobilenetv2}
M.~Sandler, A.~Howard, M.~Zhu, A.~Zhmoginov, and L.-C. Chen, ``Mobilenetv2: Inverted residuals and linear bottlenecks,'' in \emph{Proceedings of the IEEE conference on computer vision and pattern recognition}, 2018, pp. 4510--4520.

\bibitem{he2016deep}
K.~He, X.~Zhang, S.~Ren, and J.~Sun, ``Deep residual learning for image recognition,'' in \emph{Proceedings of the IEEE conference on computer vision and pattern recognition}, 2016, pp. 770--778.

\bibitem{shu2021channel}
C.~Shu, Y.~Liu, J.~Gao, Z.~Yan, and C.~Shen, ``Channel-wise knowledge distillation for dense prediction,'' in \emph{Proceedings of the IEEE/CVF International Conference on Computer Vision}, 2021, pp. 5311--5320.

\bibitem{vitmatte}
J.~Yao, X.~Wang, S.~Yang, and B.~Wang, ``Vitmatte: Boosting image matting with pretrained plain vision transformers,'' \emph{arXiv preprint arXiv:2305.15272}, 2023.

\bibitem{shen2023x}
K.~Shen, C.~Guo, M.~Kaufmann, J.~J. Zarate, J.~Valentin, J.~Song, and O.~Hilliges, ``X-avatar: Expressive human avatars,'' in \emph{Proceedings of the IEEE/CVF Conference on Computer Vision and Pattern Recognition}, 2023, pp. 16\,911--16\,921.

\bibitem{Chen_2022_ACCV}
X.~Chen, Y.~Zhu, Y.~Li, B.~Fu, L.~Sun, Y.~Shan, and S.~Liu, ``Robust human matting via semantic guidance,'' in \emph{Proceedings of the Asian Conference on Computer Vision (ACCV)}, 2022, pp. 2984--2999.

\bibitem{migacz20178}
S.~Migacz, ``8-bit inference with tensorrt,'' in \emph{GPU technology conference}, 2017, p.~5.

\end{thebibliography}
\bibliographystyle{IEEEtran}

\end{document}